\documentclass[runningheads]{llncs}
\usepackage[T1]{fontenc}
\usepackage{graphicx}
\usepackage{hyperref} 
\usepackage{color}

\usepackage{caption}
\usepackage{subcaption}
\usepackage{amsmath}

\usepackage{tabularx}
\usepackage{float}
\usepackage{algpseudocode}
\usepackage{algorithm}
\usepackage{amsfonts}
\usepackage{nicefrac}
\usepackage{booktabs}
\usepackage{microtype}      
\usepackage{xcolor}      

\algnewcommand\algorithmicforeach{\textbf{for each}}
\algdef{S}[FOR]{ForEach}[1]{\algorithmicforeach\ #1\ \algorithmicdo}

\begin{document}
\title{Group-Agent Reinforcement Learning}
%
%
\author{Kaiyue Wu \and Xiao-Jun Zeng}
\authorrunning{Kaiyue Wu \and Xiao-Jun Zeng}
%
\institute{The University of Manchester, Manchester M13 9PL, UK\\
\email{\{kaiyue.wu,x.zeng\}@manchester.ac.uk}}
\maketitle              
\begin{abstract}
It can largely benefit the reinforcement learning (RL) process of each agent if multiple geographically distributed agents perform their separate RL tasks cooperatively. Different from multi-agent reinforcement learning (MARL) where multiple agents are in a common environment and should learn to cooperate or compete with each other, in this case each agent has its separate environment and only communicates with others to share knowledge without any cooperative or competitive behaviour as a learning outcome. In fact, this scenario exists widely in real life whose concept can be utilised in many applications, but is not well understood yet and not well formulated. As the first effort, we propose group-agent system for RL as a formulation of this scenario and the third type of RL system with respect to single-agent and multi-agent systems. We then propose a distributed RL framework called DDAL (Decentralised Distributed Asynchronous Learning) designed for group-agent reinforcement learning (GARL). We show through experiments that DDAL achieved desirable performance with very stable training and has good scalability.

\end{abstract}
 \section{Introduction}
\label{intro}

Currently reinforcement learning (RL) problems are considered in two types of systems, which are single-agent system and multi-agent system. For single-agent RL problems, there is only one intelligent agent involved in the learning process. It learns through interacting with its surroundings in order to achieve the objective of optimal individual behaviour. For multi-agent RL problems, there are multiple agents involved in the learning process. They can learn not only through interacting with their surroundings but also through interacting with each other, with the objective of not only optimal individual behaviour but also optimal team behaviour of cooperation or equilibrium behaviour of competition. These two parts of the learning objective are not separable since cooperative or competitive behaviour is actually an essential part of individual behaviour in multi-agent reinforcement learning (MARL). 

However, there are some real-world RL scenarios which cannot be classified into either of the two categories and still lack understanding. In existing literature, there is a lack of understanding for pure cooperative learning and an ambiguity between cooperative learning and learning to cooperate. In a typical multi-agent problem, such as robotic soccer playing, all the agents perform learning activities together in a common environment. For each single agent, the other agents are part of its environment which becomes non-stationary due to the continually changing behaviour of those agents \cite{DBLP:journals/corr/abs-2011-11012}. In this case, all agents from one team are learning together to cooperate with each other to achieve a common goal. Obviously learning to cooperate often involves cooperative learning (completely independent learning is also possible) which is the learning process with knowledge shared among agents, since agents would probably need information from others to cooperate with them. But from the other way around, cooperative learning does not involve learning to cooperate very often. Agents can be learning with different goals in their own separate environment which is not affected by others and purely communicating knowledge to benefit each other's learning process without any cooperative or competitive behaviour as a learning outcome. The only cooperation is the communication during learning process, which is native and not learned behaviour. We name this learning system a group-agent system that aims at connecting learning agents that are naturally geographically distributed for the purpose of improving learning behaviour. 

This scenario is fundamentally different from MARL since it is doing pure cooperative learning for the purpose of acquiring only optimal individual behaviour (potentially diversified and not containing cooperative or competitive behaviour) faster and better while MARL is learning to cooperate or compete. And it certainly cannot be categorised as single-agent RL since it involves a group of agents with separate tasks. In another word, it identifies a gap which has not been addressed by traditional RL framework.
A simple real-life example would be multiple students study by themselves in different countries through environment interactions while also communicate with each other online in a "study group" to share knowledge. These countries are geographically distributed thus not affected by each other, and Internet serves as the communication channel between them. We can see from the example that via this pure cooperative learning each agent is able to learn faster to achieve its goal in its own environment. It can be useful in many applications, such as autonomous driving and networking, which motivates us to formulate it as group-agent reinforcement learning (GARL) to inspire dedicated approaches to solve it. In this paper, we propose a cooperative distributed RL framework called DDAL (Decentralised Distributed Asynchronous Learning) as a proof-of-concept effort to tackle GARL. The empirical evaluation shows DDAL achieved good performance with remarkable training stability.

\section{Background}
\label{back}

RL is a process of learning by trial and error. In this process, there is one or several intelligent agents interacting with their surroundings from which they could get feedbacks for the actions they take. In this way, the agents are able to obtain knowledge on how to behave better and gradually improve their performance. 
Single-agent RL is often modeled as a Markov Decision Process (MDP) \cite{10.2307/24900506}. All the states satisfy the memoryless state transition property $P[S_{t+1} | S_t] = P[S_{t+1} | S_1, S_2, ... , S_t]$ where the next state is only relevant to the current state without being affected by any previous states or we can say that the current state grabs all necessary information from past states. An MDP is basically a tuple
\begin{equation}
<\mathcal{S}, \mathcal{A}, \mathcal{P}, \mathcal{R}, \gamma>
\end{equation}
where $\mathcal{S}$ is a finite set of environment states, $\mathcal{A}$ is a finite set of actions, $\mathcal{P}$ is the state transition probability matrix where $P_{s,s'} = P[S_{t+1}=s' | S_t=s]$ is the probability of transiting from state $s$ to $s'$, $\mathcal{R}$ is reward function, and $\gamma$ is the discount factor used in the Bellman equation formulation of value functions. For example, the state value function $V(s)$ can be stated as $V(s) = E[R_{t+1} + \gamma V (S_{t+1})|S_t = s]$. 
Given the action selection policy $\pi(a|s)$ which is a probability distribution over all possible actions under state $s$, we can further have $P[S_{t+1}=s' | S_t=s] = \sum_a P[S_{t+1}=s'|S_t=s,A_t=a] \cdot \pi(a|s)$. 

Multi-agent reinforcement learning is often modeled as a stochastic game which is a generalisation of MDP to the multi-agent case \cite{bucsoniu2010multi}. It is stated by the following tuple
\begin{equation}
<\mathcal{S}, \mathcal{A}_1, \cdots, \mathcal{A}_n, \mathcal{P}, \mathcal{R}_1, \cdots, \mathcal{R}_n, \gamma>
\end{equation}
where $n$ is the number of agents, $\mathcal{S}$ is a finite set of environment states, $\mathcal{A}_i, i=1,\cdots,n$ are the finite sets of actions of every agent, $\mathcal{P}$ is the state transition probability matrix, $\mathcal{R}_i, i=1,\cdots,n$ are the reward functions of every agent, and $\gamma$ is the discount factor. From this formulation, we can see that all the agents are in a common environment so that we have only one set of environment states, and the state transition is the same from every agent's point of view. 

\section{Related Work}
\label{re}

Methods for single-agent RL can be classified into two broad categories, basic algorithms and distributed variants (parallelisation) of the basic algorithms. For basic algorithms, there are the well-known Q-learning introduced by \cite{watkins1989learning} which maintains and updates a Q-table during training. 
Replacing the Q tables with deep neural networks, called Q networks, we have DQN (Deep Q-Network) \cite{cite-key} which takes raw RGB images as input and is trained to output the Q values for all possible state-action pairs $(s,a_i)$. 
Other than the value-based methods, we also have actor-critic track. A2C, short for Advantage Actor-Critic, is a typical kind of actor-critic method for RL. It has two neural networks for approximating policy function $\pi_{\theta}$ and state-value function $V(s)$. 
Another popular actor-critic method is PPO \cite{schulman2017proximal}, short for Proximal Policy Optimisation, which introduces an innovative clipped surrogate loss for the policy function. 

For the distributed variants, there are A3C (asynchronous advantage actor-critic) \cite{mnih2016asynchronous} which is an asynchronous version of distributed A2C, Gorila \cite{nair2015massively} which is distributed DQN, and APPO \cite{luo2020impact} which is asynchronous PPO. A3C has a central copy of network models asynchronously receiving gradients from multiple parallel A2C workers and periodically synchronises the workers with the central copy. 
Gorila has multiple learner processes training a central Q-network copy with experiences generated by multiple parallel actor processes interacting with environment. 
APPO has a PPO learner paired with multiple actor processes who generate experiences. The experiences are stored in a circular buffer and will be discarded after being used in training. 
These distributed RL algorithms are single-agent parallelisation and cannot be applied to GARL. But each agent in GARL should be able to apply any one of the single-agent algorithms. DDPPO \cite{wijmans2020ddppo}, short for decentralised distributed PPO, is another parallel single-agent algorithm, but is a bit different in that it involves multiple "agents", respects the completeness of each "agent" and does not parallelise the internal processes of an agent such as actor or learner processes as the other distributed algorithms do. It is completely decentralised without any central network copy. All the "agents" update their models locally after communicating gradients directly with each other. However, it does global synchronous control among the "agents" -- the communication and model updates all happen synchronously with all updates that happen at the same time being identical. This breaks the autonomy of the "agents" and actually makes them worker copies of one single agent. Hence it is still a single-agent learning system, but when GARL agents all work on same tasks we can say that GARL is generalisation of it that additionally respects agent autonomy. We will compare our proposed method DDAL with the synchronous method behind DDPPO in section \ref{experiments}.

Methods for multi-agent reinforcement learning can be classified into three broad categories, cooperative algorithms, competitive algorithms and algorithms for a mix of cooperation and competition. 
The competition or cooperation here refers to the interactions between agents which is learned behaviour and the nature of the problem, in comparison to cooperative learning where the cooperation is only knowledge communication and not learned behaviour. MARL methods do not apply in GARL due to the inherent difference in problem definition. To justify more, multi-agent problem can be approached through methods with only independent learner which optimises its own policy ignoring the other agents and assuming environment stationarity \cite{matignon2012independent,foerster2017stabilising}. More approaches would consider environment non-stationarity and study the joint behaviour of the agents instead \cite{zhang2018networked,wang2020qplex,ma2021modeling}. For example, \cite{lowe2017multi} considers a centralised critic that takes the states and actions of all agents and outputs the Q value for each agent, while also maintains approximated policies of other agents at each agent. 
In GARL, the agents share knowledge with each other (thus not independent) only to benefit each other's learning (thus joint behaviour optimisation is not necessary). Besides, MARL happens in a common environment thus the environment state is identical to every agent at any time point, while in GARL the environment states are varied (the agents can be at different pace even though with same tasks). Considering this variousness is an important task of GARL methods. And the multiple geo-distributed environments in GARL also introduce many issues from the systems side. In real applications geo-distributed agents would need real communication and subsequently be managed by a networking protocol, resulting in a real system rather than just a machine learning algorithm. 

We also notice that multi-task learning is trying to learn a single policy that works across a set of related tasks within the same environment \cite{vithayathil2020survey}. It can be studied in either single-agent system or multi-agent system while the nature of the multi-tasking problem is not affected. An agent learns for a generalised policy that works across a set of related tasks in its own single environment or within the common environment \cite{omidshafiei2017deep}. 
This diverges from our focus of studying the group learning behaviour where multiple autonomous agents separately learn in their own environment while communicate with each other to benefit each other's learning process that only focuses on its own task, which will result in different policies with possibly different expertise among the agents. 

\section{Group-Agent Reinforcement Learning (GARL)}
\label{group}


In GARL, there are multiple agents doing RL together in a "study group", which is abstracted from a very common real-life behaviour in human intelligence. When we humans study, there are basically two knowledge sources, learning through trial and error in our environment (RL) and learning cooperatively through retrieving available knowledge from other people. Hence we often study together in groups to benefit the latter process. 
It does not have to happen in a single environment, but can rather work across multiple environments. GARL agents learn through trial and error in their separate environment while communicating with each other to obtain available knowledge. Each of these environments is stationary because no one will interfere with others' environment. From another perspective, what GARL does is connecting distributed autonomous learning agents for them to share knowledge, leveraging the power of the learning community. 


We take autonomous driving as an example application to give further explanation. The training of self-driving cars can well take place with RL \cite{sallab2017deep}. We describe it through three training stages, where stage 2 is an example of GARL. 

\begin{itemize}
\item Stage 1: Given a certain city environment, one single self-driving car is doing RL to obtain driving knowledge in one neighbourhood. Its environment, namely this neighbourhood, is stationary. Learning only happens through trial and error. 
\item Stage 2: Still in the same city environment, there are now multiple self-driving cars all doing RL simultaneously, each in a different neighbourhood. Each of their environments, namely the neighbourhoods, is still stationary. The goal of every agent is to learn to drive in its own neighbourhood environment. We can see that the goals among the agents are slightly different due to the difference between the neighbourhoods. However since these neighbourhoods belong to the same city environment, they share much similarity. Therefore, it will largely benefit learning if we create communication channels between the agents for them to exchange their knowledge acquired through environment exploration. It is very possible that one car is not able to explore its environment thoroughly and leave out many environment states, but some other peer car explores them well, so that sound knowledge can be obtained through communicating with that peer car. In this case, learning happens in a group-agent setting. 
\item Stage 3: With the help of GARL, the multiple self-driving cars all learned to drive in its own environment well and fast. Now some of the cars drive out of their neighbourhoods to meet other peer cars. In one neighbourhood, there are several cars on the road. They need to learn to cooperate with each other to safely co-exist on the road, not causing any car crashes. This neighbourhood environment becomes non-stationary since each of these cars becomes a part of the others' environment and their behaviours are continually evolving. This turns to be an MARL scenario. 
\end{itemize}

Note that GARL cannot be viewed as a simplified version of MARL with just the objective of cooperation or competition gotten rid of. In GARL, since each agent is in a separate environment, they can have different state sets and diversified individual learning goals. There is inherently much more freedom for the agents compared to MARL where the agents are very restricted by each other. We present this more clearly with a formal formulation as follows.


Recall from section \ref{back} that single-agent RL can be modeled as an MDP and MARL can be modeled as a stochastic game. Here we propose group MDP to state GARL, in the following tuple
\begin{equation}
\begin{aligned}
<\mathcal{S}_1, \cdots, \mathcal{S}_n, \mathcal{A}_1, \cdots, \mathcal{A}_n, \mathcal{P}_1, \cdots, \mathcal{P}_n, \mathcal{R}_1, \cdots, \mathcal{R}_n, \\
\gamma_1, \cdots, \gamma_n, \mathcal{K}_1, \cdots, \mathcal{K}_n, \mathcal{K}_{-1}, \cdots, \mathcal{K}_{-n}>
\end{aligned}
\end{equation}
where $n$ is the number of agents, $\mathcal{S}_i, \mathcal{A}_i, \mathcal{P}_i, \mathcal{R}_i, \gamma_i, \mathcal{K}_i, \mathcal{K}_{-i}, i=1,\cdots,n$ are the sets of environment states, the sets of actions, the state transition probability matrixes, the reward functions, the discount factors, the sets of knowledge from local environment interactions and the sets of received knowledge of every agent in the group. Note that $\mathcal{K}_{-i} = \{\mathcal{K}_{1,i}, \cdots, \mathcal{K}_{i-1,i}, \mathcal{K}_{i+1,i}, \cdots, \mathcal{K}_{n,i}\}$ where $\mathcal{K}_{i,i'} \subseteq \mathcal{K}_i$ is the knowledge of agent $i$ shared to agent $i'$, and for at least one pair of $i, j \in \{1, \dots, n\}$, $\mathcal{S}_i \cap \mathcal{S}_j \neq \emptyset$. Each agent can send its knowledge to any other agents arbitrarily and store its received knowledge in local memory for training. From this formulation, we can see that different from MARL where the state set and state transition probability matrix are shared among all agents, each agent in GARL works in its own separate environment so that it has its own set of environment states, and an agent's environment is independent of any other agent's environment so that it has its own state transition probability matrix. Every agent has its own set of actions, reward function, discount factor, set of local knowledge and set of received knowledge. 
Note that the knowledge shared among agents can be in various forms, such as raw experiences (state, action, reward tuple), policy parameters, state/action values, gradients at each update iteration, etc. With this knowledge sharing, GARL aims to benefit each single agent's learning quality and speed. 
It can be applied to the training of video game playing and autonomous driving. Besides, we claim that it has great potential in network routing problems due to the independent environment of each network node and the natural communication network between them. 

\section{Decentralised Distributed Asynchronous Learning (DDAL)}
\label{DDAL}

This section introduces DDAL as a proof-of-concept learning framework designed for GARL. The idea is fourfold:

\begin{itemize}
\item Decentralised control: The group-agent system naturally comes in a decentralised manner where every agent is autonomous and can learn independently. Artificially having them managed by a central controller can be expensive and sometimes meaningless or unrealistic. Thus we apply decentralised control.
\item Asynchronous communication: To give as much freedom as possible to the agents and respect their nature of autonomy, we design to let the communication happen in an asynchronous manner. Synchronous communication among distributed autonomous agents means that the agents should all agree to dedicated communication stages when they are all sending or receiving messages. This can be very difficult in real-world applications since organising these autonomous agents needs lots of efforts from the perspective of distributed systems. Thus we apply asynchronous communication where each agent can send knowledge to other agents or receive knowledge from them at any convenient time, avoiding the need for a communication protocol.
\item Independent learning at beginning stage: To explain this from intuition, we are probably not able to acquire very accurate knowledge at the beginning stage of our learning by trial-and-error (due to the inaccurate measurement of error under insufficient prior knowledge) and sharing of beginners' mistakes would have negative effect on others' learning processes, hence it is good practice to start group communication after everyone has reached a relatively stable learning status.
\item Weighted gradient average: Here we use gradients as the form of knowledge among agents and require that all gradients ever generated will be shared to every other agent ($\mathcal{K}_{i,i'} = \mathcal{K}_i$, $i,i'=1,\cdots,n$). 
Each piece of gradients (for one model update) from any agent is accompanied with two extra pieces of information, the learning experience so far and its relevance to the agent that it's going to. For example, the number of training epochs performed by an agent can represent the learning that the agent has experienced so far, namely the amount of training so far for the piece of gradients just generated by this agent. We quantify these two pieces of information with $T_j$ (training experience) and $R_j$ (relevance) for the $j-th$ piece of gradients represented as $g_j$ in a chunk of received gradient pieces. When agent $i$ is ready to perform a model update involving received gradients, it retrieves $m$ pieces of gradients from $\mathcal{K}_{i} \cup \mathcal{K}_{-i}$ and calculates a weighted gradient average according to the equation $\overline{g} = \frac{1}{2} ( \sum_{j=1}^{m} \frac{T_j}{\sum_{j=1}^{m} T_j} g_j + \sum_{j=1}^{m} \frac{R_j}{\sum_{j=1}^{m} R_j} g_j )$, 
then perform the update with $\overline{g}$. The average operation allows us to mitigate the influence introduced by poor experiences and introducing weights lowers the influence of immature or irrelevant knowledge.
\end{itemize}

The algorithm at each agent is shown in Algorithm \ref{alg1}. After being trained for a number of epochs, the agent starts to send its gradients to other agents and perform model updates with received gradients every few epochs. The threshold and minibatch size are hyper parameters. We do not have global organising mechanism for the agent system thus each agent is basically on its own. In our implementation, this decentralised control and asynchronous communication is realised through multiprocessing queues. Every agent has its own queue to hold the knowledge received from other agents, and these queues are shared among all agents so that each agent is free to send its knowledge to any other agent's queue. The agents are implemented with Salina \cite{salina}. 
We claim that DDAL should not be restricted by agent type. The single-agent algorithms as discussed in section \ref{re} should all be able to serve as our agent's brain. Here we discuss a classic A2C agent.

\begin{algorithm}
\caption{DDAL at the $i-th$ agent} \label{alg1}
\begin{algorithmic}[1]
\Require Initialise knowledge set $\mathcal{K}_i$ and $\mathcal{K}_{-i}$
\ForEach{$epoch$}
\State Generate $k$ experiences
\State Compute average loss
\State Compute gradients
\If{$epoch < threshold$}
\State Update model with the gradients
\Else
\State Append the gradients with weighting information $T$ and $R$
\State Store the gradients in $\mathcal{K}_i$
\State Send a copy of the gradients to every other agent $j$ (stored in $\mathcal{K}_{-j}$) ($j=1,\cdots,i-1,i+1,\cdots,n$)
\If{$epoch \% minibatch == 0$}
\State Get (and remove) $m$ pieces of gradients from $\mathcal{K}_i \cup \mathcal{K}_{-i}$
\State Compute $\overline{g}$ of these gradients
\State Update model with $\overline{g}$
\EndIf
\EndIf
\EndFor 
\end{algorithmic}
\end{algorithm}


  
\subsection{Decentralised Distributed Asynchronous Advantage Actor-Critic (DDA3C)}

For a classic A2C agent, with the relation $Q(s_t,a_t) = A(s_t,a_t) + V(s_t)$ where $A(s_t,a_t)$ is the advantage value, 
we have the gradients for policy network as $\nabla_{\theta} log \pi_{\theta}(a_t|s_t)A(s_t,a_t) = \nabla_{\theta} log \pi_{\theta}(a_t|s_t)(Q(s_t,a_t) - V(s_t))$ where $Q(s_t,a_t) = r + \gamma V(s_{t+1})$ (=r, for terminal $s_{t+1}$). 
With this A2C agent, we name the complete algorithm as DDA3C. 

\section{Evaluations}
\label{experiments}

\begin{figure*}
        \centering
        \begin{subfigure}[b]{0.33\textwidth}
                \centering
                \includegraphics[width=\linewidth]{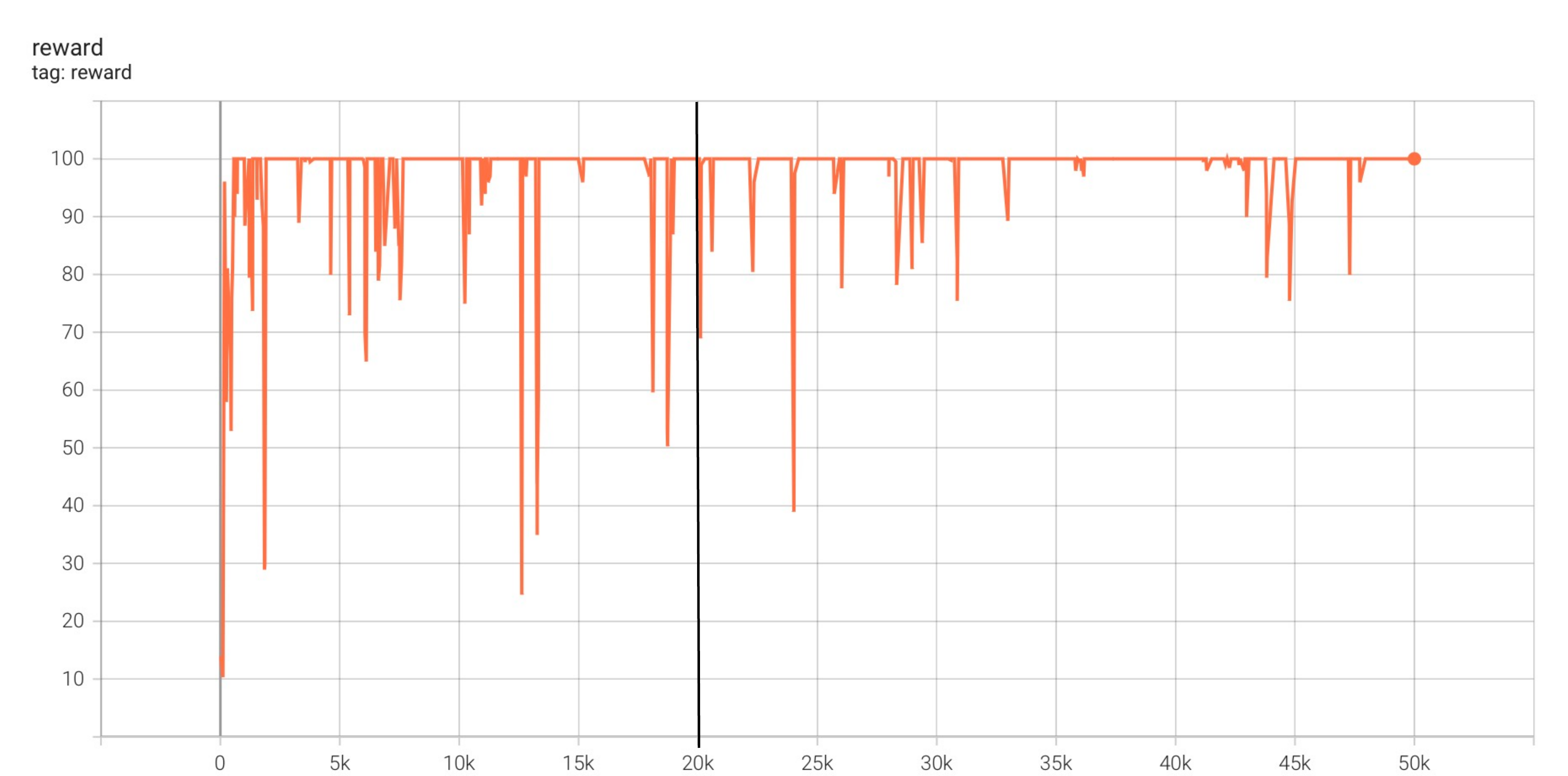}
                \caption{Single-agent}
                \label{fig:1}
        \end{subfigure}%
        \begin{subfigure}[b]{0.33\textwidth}
                \includegraphics[width=\linewidth]{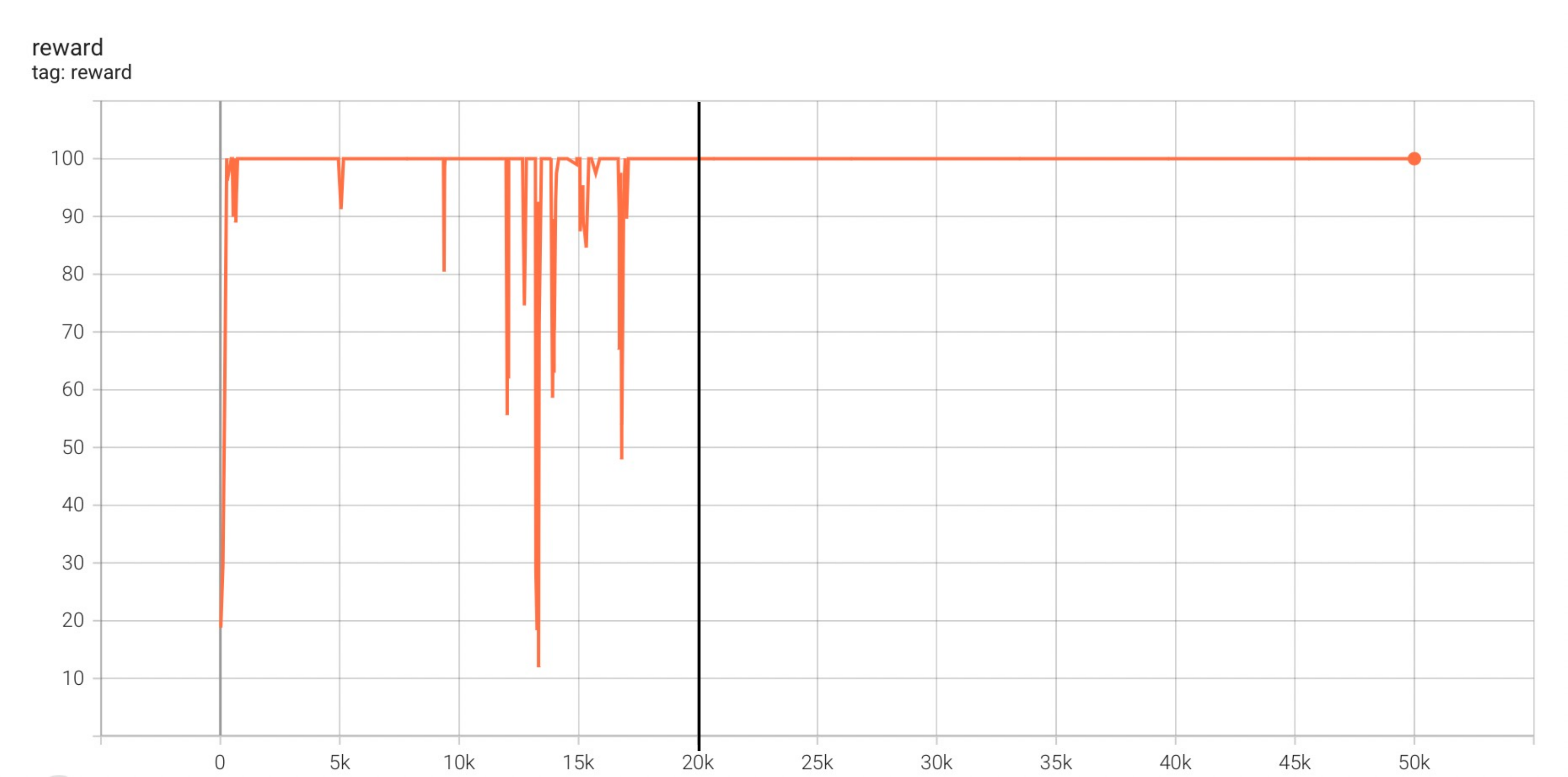}
                \caption{Group-agent: agent 1}
                \label{fig:2}
        \end{subfigure}%
        \begin{subfigure}[b]{0.33\textwidth}
                \includegraphics[width=\linewidth]{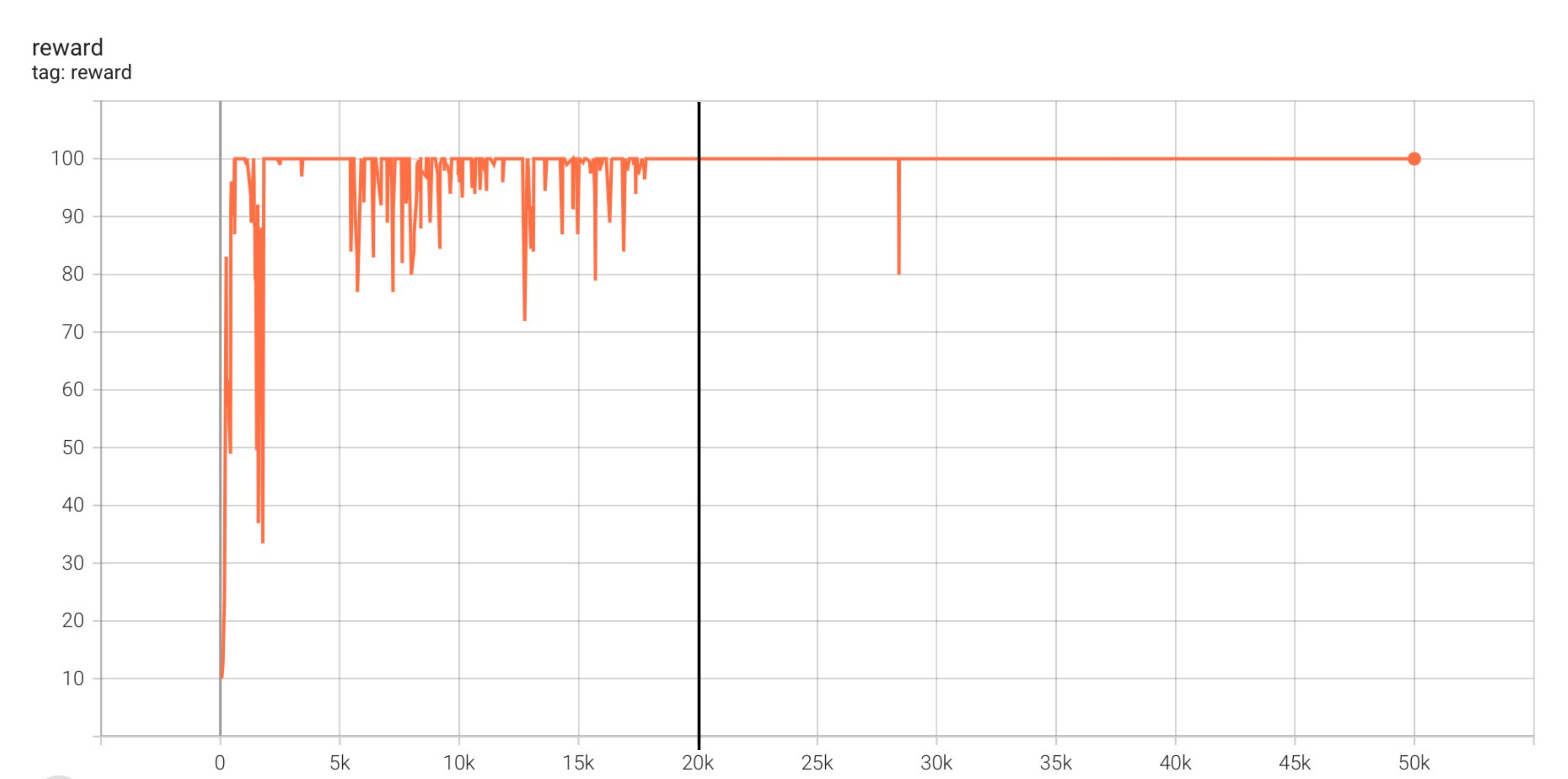}
                \caption{Group-agent: agent 2}
                \label{fig:3}
        \end{subfigure}%
        \caption{DDA3C single-agent vs. group-agent (2 agents)}\label{fig:animals}
\end{figure*}

\begin{figure*}
         \begin{subfigure}[b]{0.25\textwidth}
                \includegraphics[width=\linewidth]{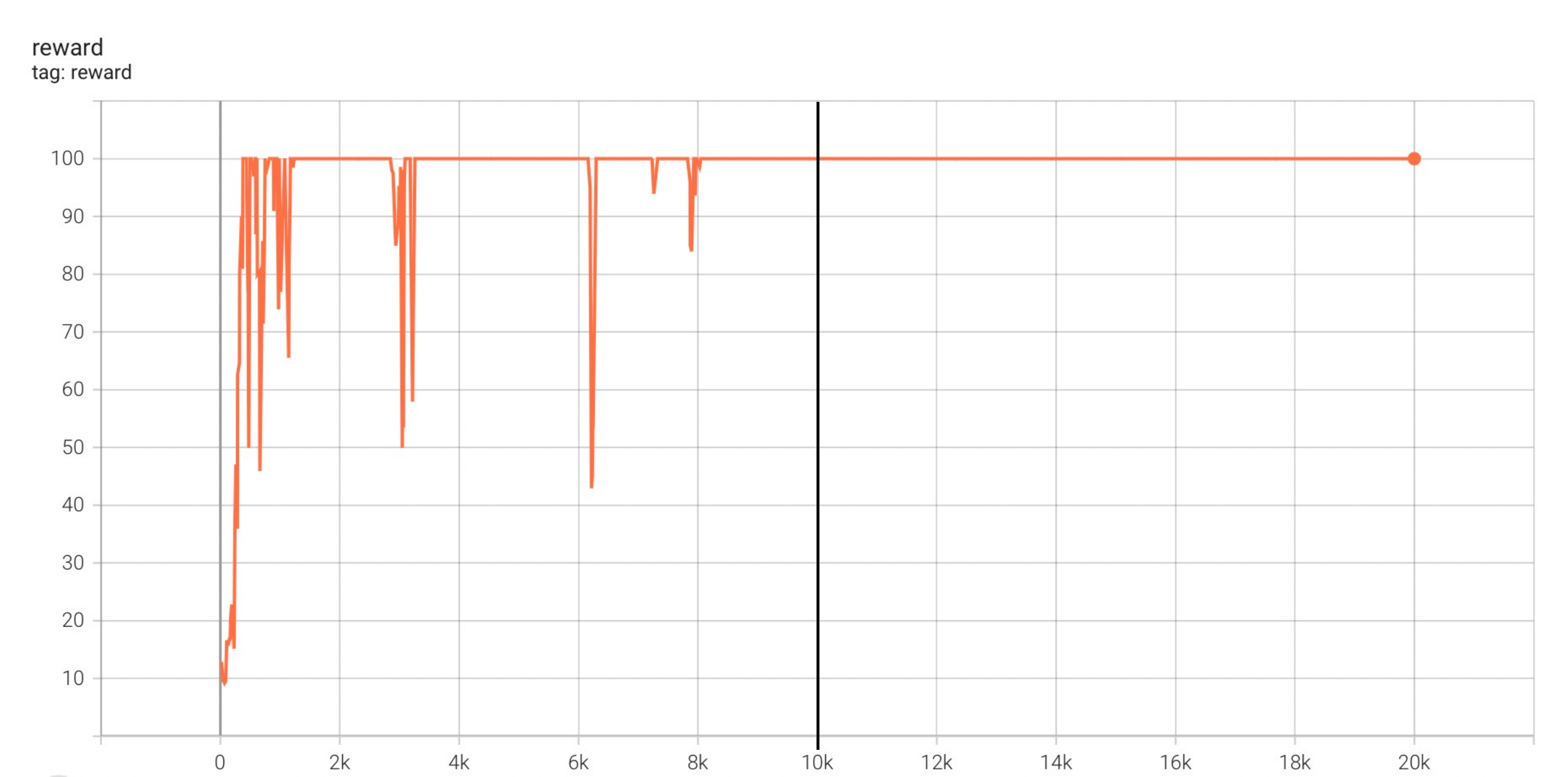}
                \caption{agent 1}
        \end{subfigure}%
        \begin{subfigure}[b]{0.25\textwidth}
                \includegraphics[width=\linewidth]{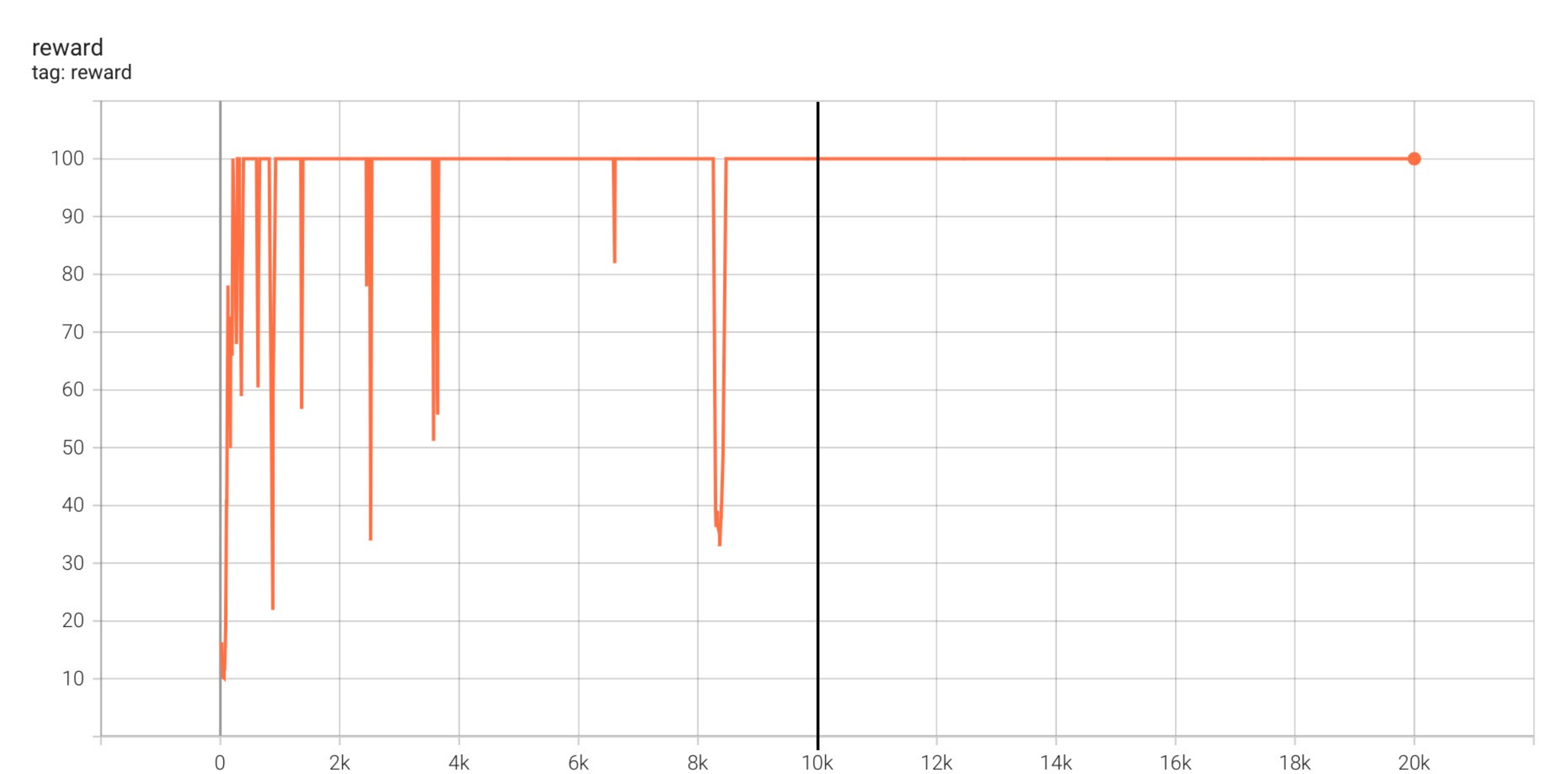}
                \caption{agent 2}
        \end{subfigure}%
        \begin{subfigure}[b]{0.25\textwidth}
                \includegraphics[width=\linewidth]{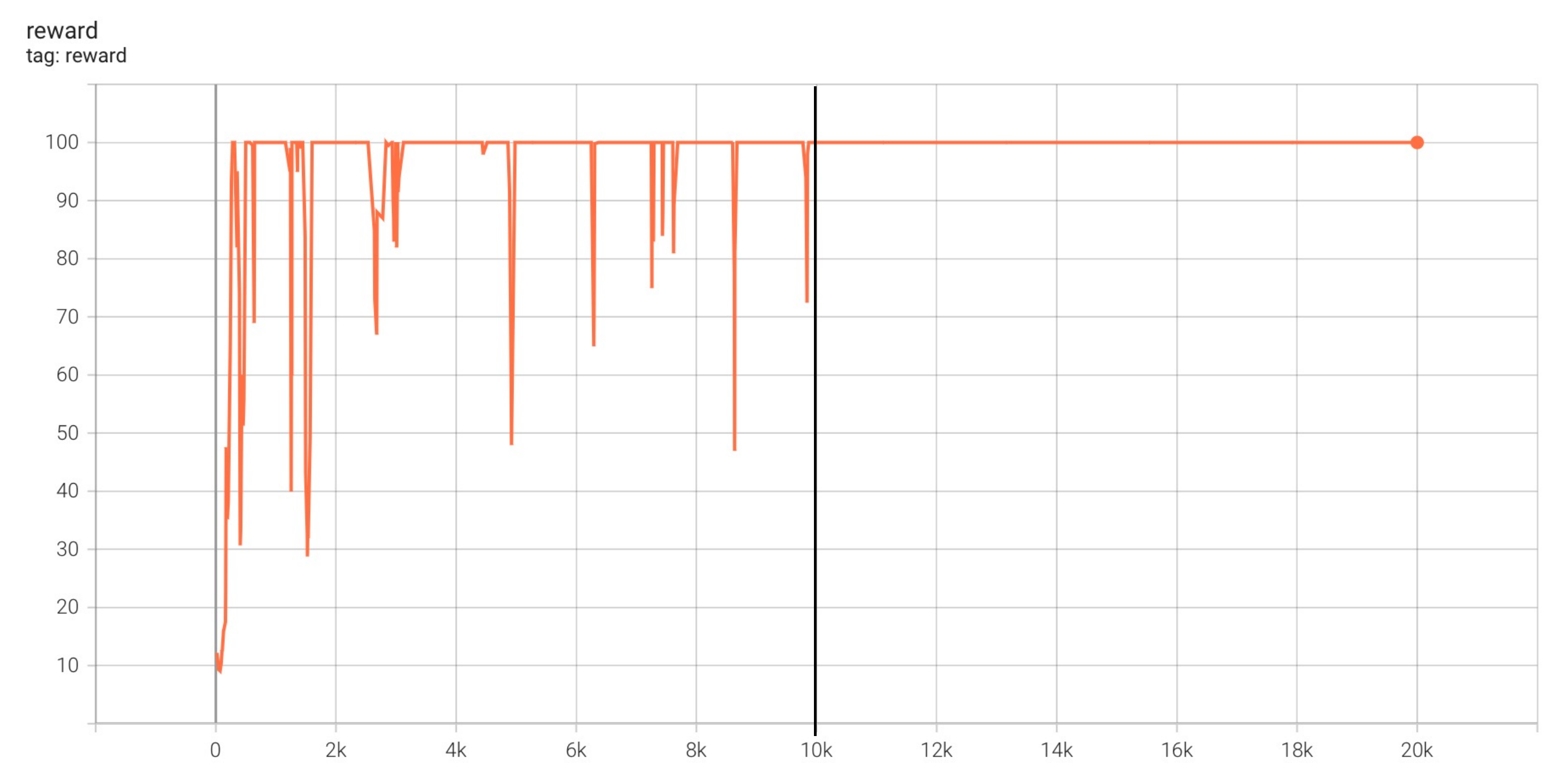}
                \caption{agent 3}
        \end{subfigure}%
        \begin{subfigure}[b]{0.25\textwidth}
                \includegraphics[width=\linewidth]{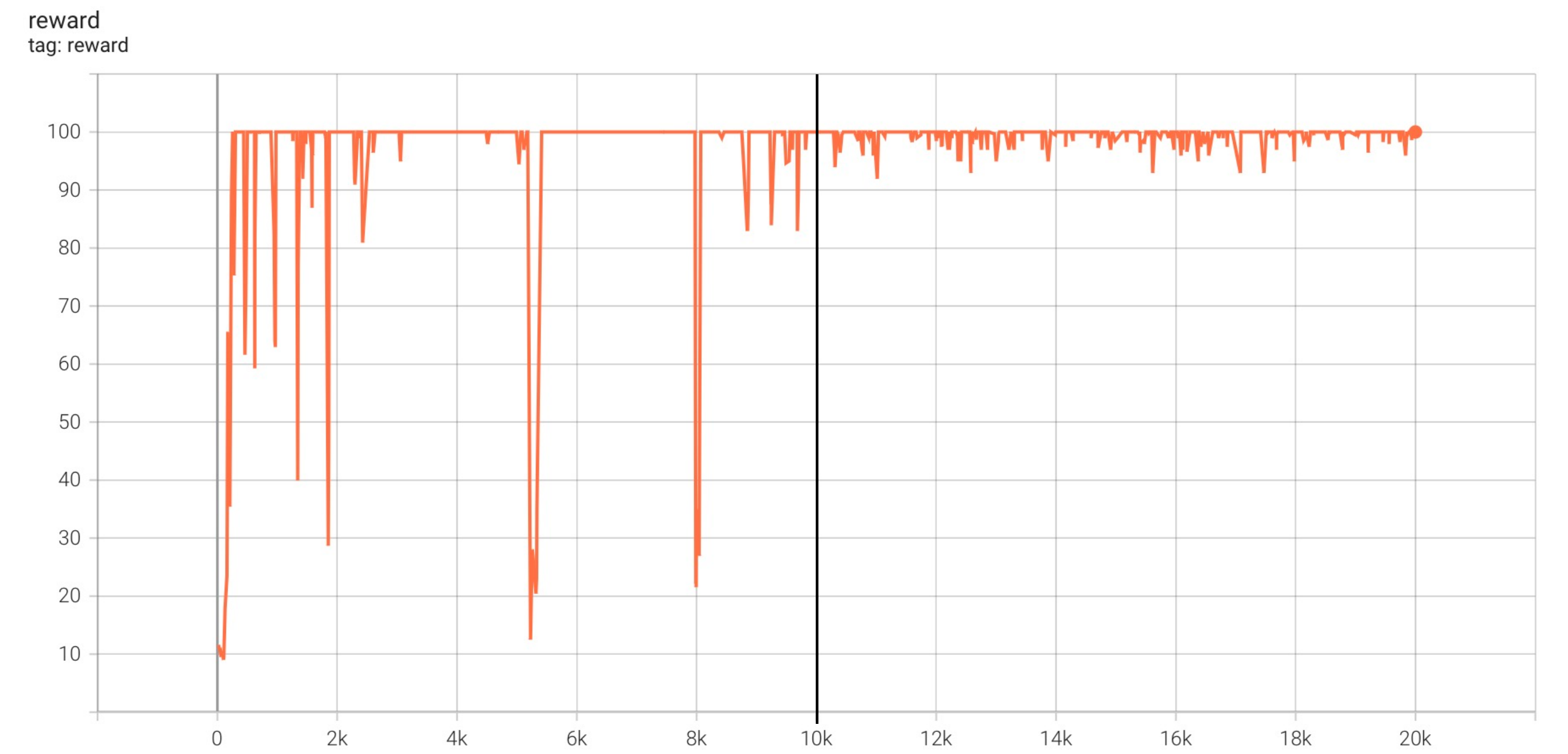}
                \caption{agent 4}
                \label{fig:tiger1}
        \end{subfigure}%
        \caption{DDA3C group-agent (4 agents)}\label{fig:4agents}
\end{figure*}

\begin{figure*}
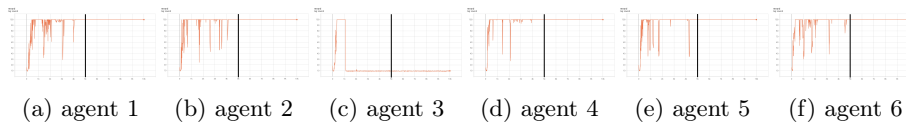

         \begin{subfigure}[b]{0.16666667\textwidth}
                \includegraphics[width=\linewidth]{6agents-1.jpg}
                \caption{agent 1}
                \label{fig:tiger2}
        \end{subfigure}%
        \begin{subfigure}[b]{0.16666667\textwidth}
                \includegraphics[width=\linewidth]{6agents-2.jpg}
                \caption{agent 2}
                \label{fig:tiger}
        \end{subfigure}%
        \begin{subfigure}[b]{0.16666667\textwidth}
                \includegraphics[width=\linewidth]{6agents-3.jpg}
                \caption{agent 3}
                \label{fig:tiger4}
        \end{subfigure}%
        \begin{subfigure}[b]{0.16666667\textwidth}
                \includegraphics[width=\linewidth]{6agents-4.jpg}
                \caption{agent 4}
                \label{fig:tiger}
        \end{subfigure}%
        \begin{subfigure}[b]{0.16666667\textwidth}
                \includegraphics[width=\linewidth]{6agents-5.jpg}
                \caption{agent 5}
                \label{fig:tiger3}
        \end{subfigure}%
        \begin{subfigure}[b]{0.16666667\textwidth}
                \includegraphics[width=\linewidth]{6agents-6.jpg}
                \caption{agent 6}
                \label{fig:tiger}
        \end{subfigure}%
        \caption{DDA3C group-agent (6 agents)}\label{fig:6agents}
\end{figure*}

In this section, we evaluate DDA3C on a scenario where there is a group of agents each of whom plays a separate instance of the same computer game while sharing knowledge with each other. Due to the consistency in learning environments and goals, every agent's knowledge is of equal relevance to other agents so that we set the $R_j$ parameters of gradients all equal to each other. And since all agents start at the same time, the $T_j$ parameters are also set identical for every piece of gradients. $m$ is the total number of gradient pieces in $\mathcal{K}_i \cup \mathcal{K}_{-i}$ at the time. Note that the result is from a single run but can represent the average performance. We have run the experiments for quite a number of times and are very confident about the result. The reason why we did not do real average performance is that for each single run the big performance fluctuations happen at different time and doing an average will seriously reduce the significance of the fluctuations. 


\subsection{DDA3C}

\begin{figure*}
         \begin{subfigure}[b]{0.125\textwidth}
                \includegraphics[width=\linewidth]{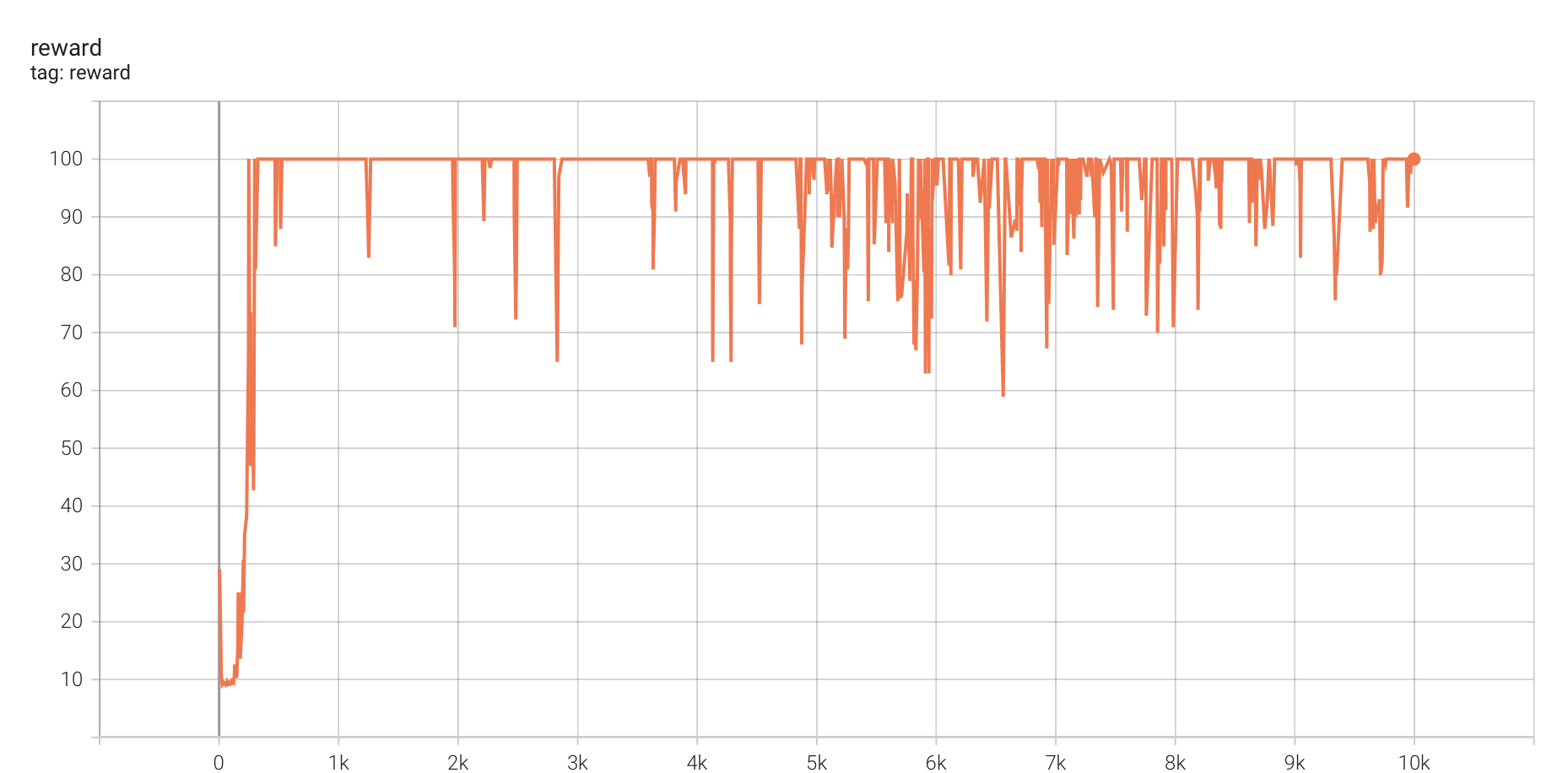}
                \caption{agent1}
                \label{fig:tiger2}
        \end{subfigure}%
        \begin{subfigure}[b]{0.125\textwidth}
                \includegraphics[width=\linewidth]{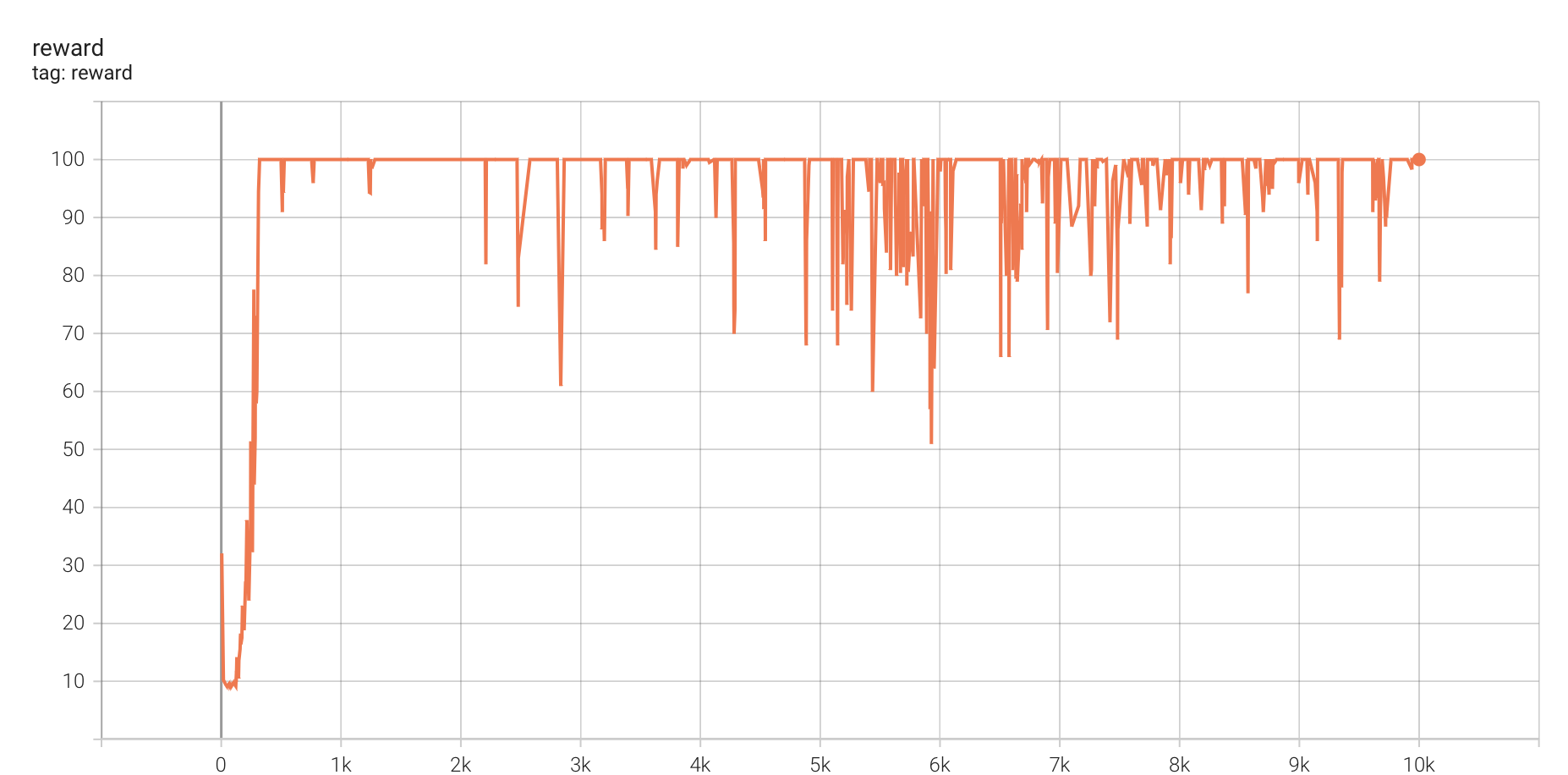}
                \caption{agent2}
                \label{fig:tiger}
        \end{subfigure}%
        \begin{subfigure}[b]{0.125\textwidth}
                \includegraphics[width=\linewidth]{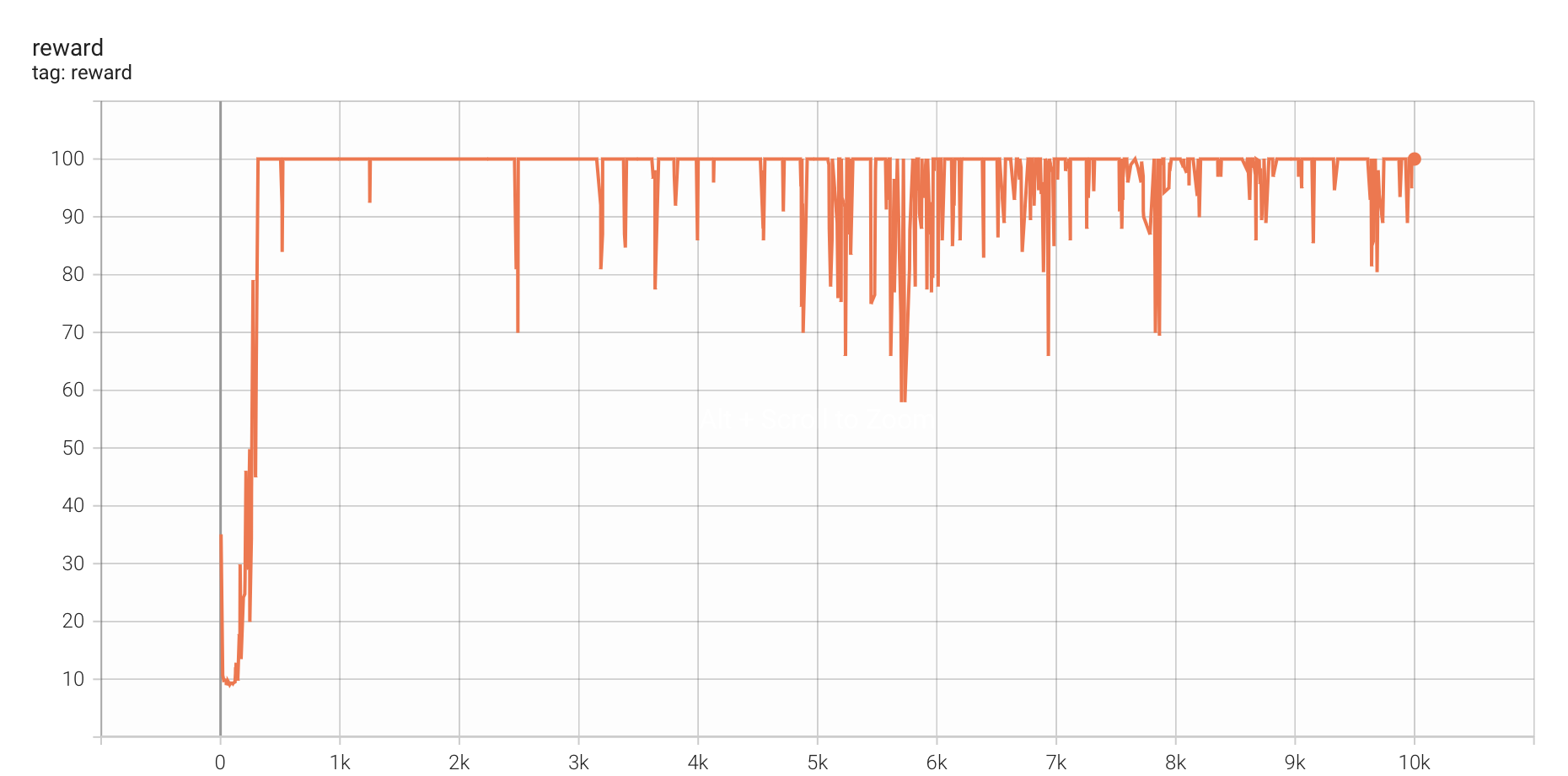}
                \caption{agent3}
        \end{subfigure}%
        \begin{subfigure}[b]{0.125\textwidth}
                \includegraphics[width=\linewidth]{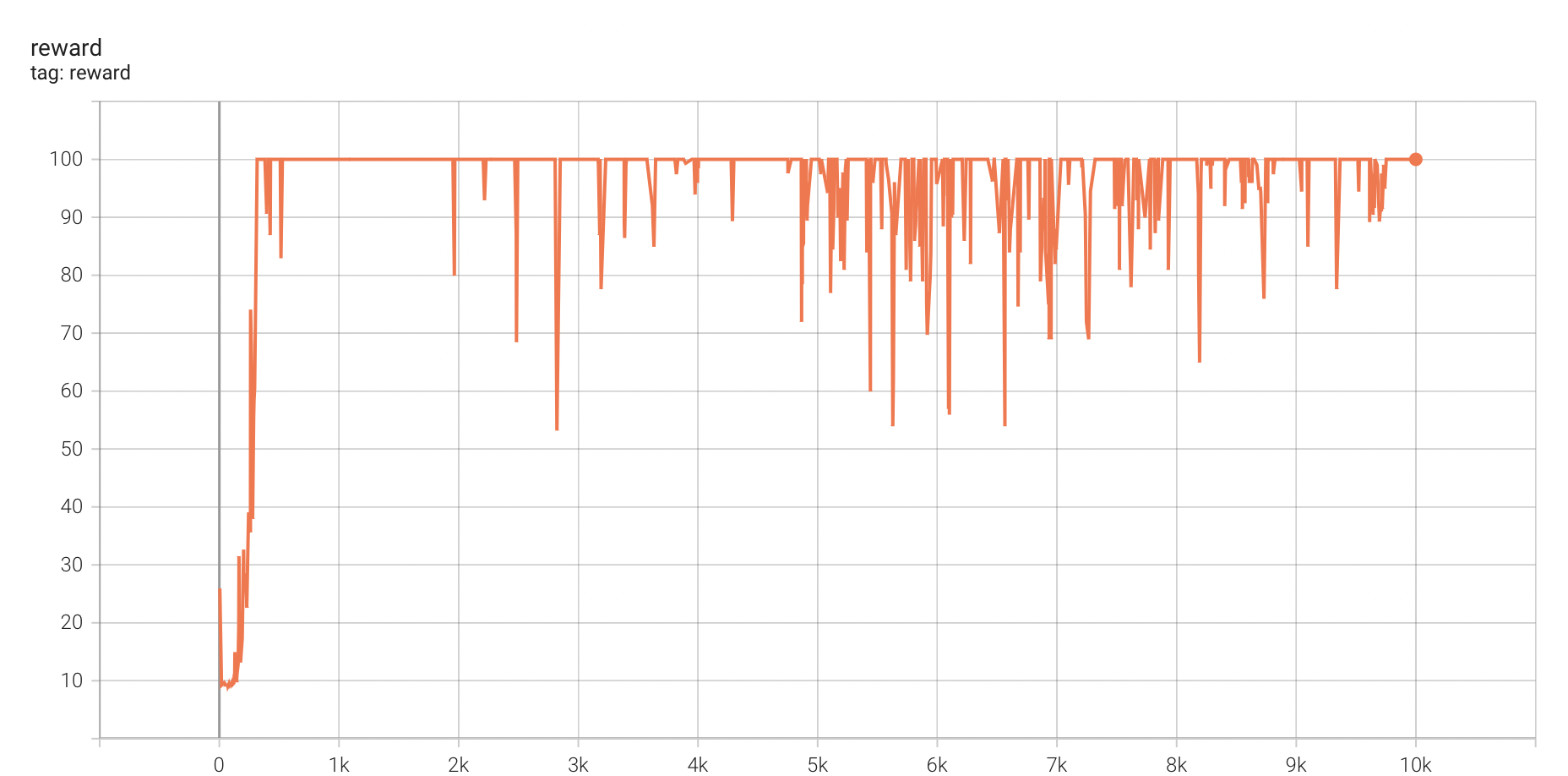}
                \caption{agent4}
                \label{fig:tiger}
        \end{subfigure}%
        \begin{subfigure}[b]{0.125\textwidth}
                \includegraphics[width=\linewidth]{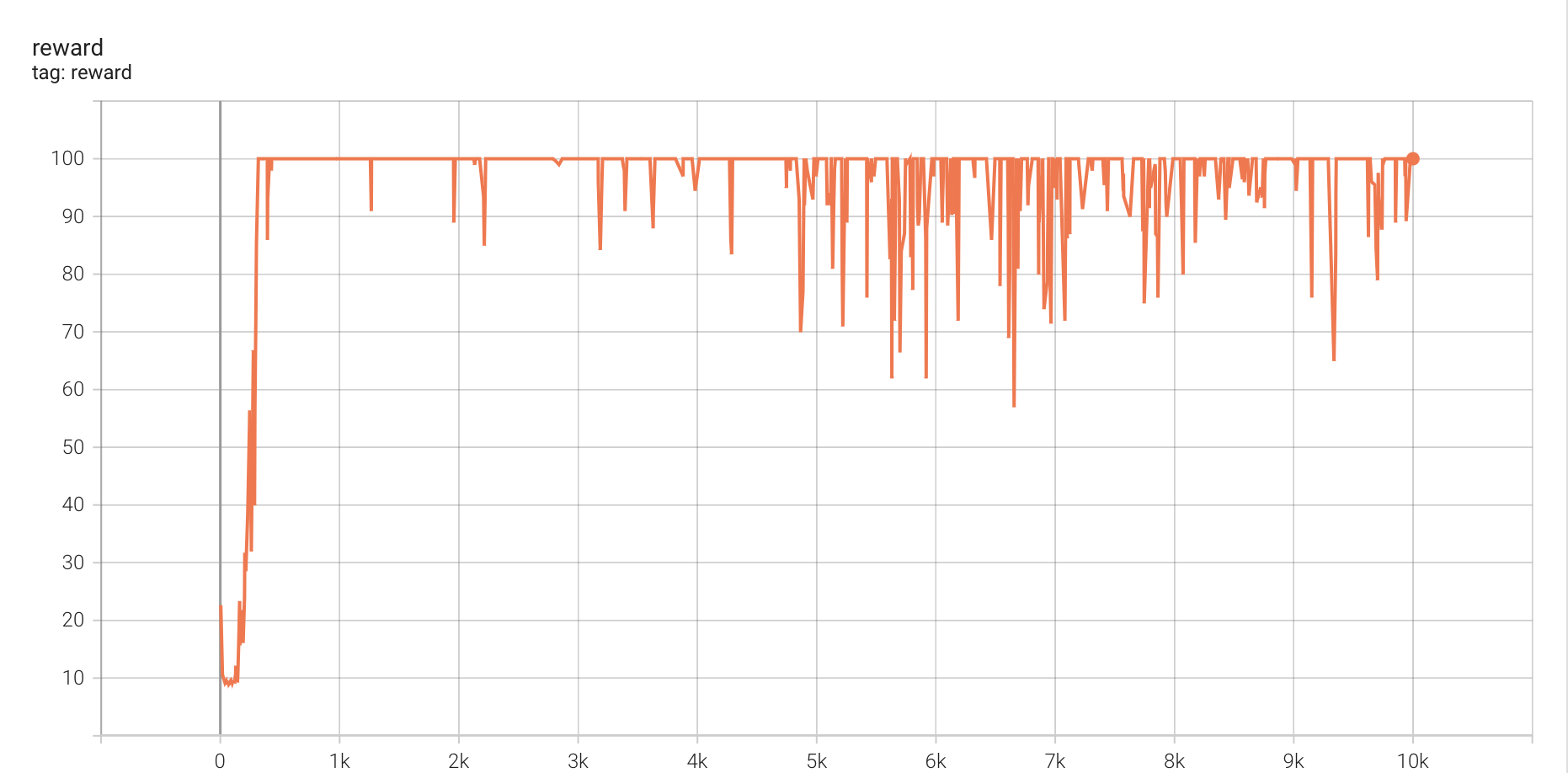}
                \caption{agent5}
                \label{fig:tiger3}
        \end{subfigure}%
        \begin{subfigure}[b]{0.125\textwidth}
                \includegraphics[width=\linewidth]{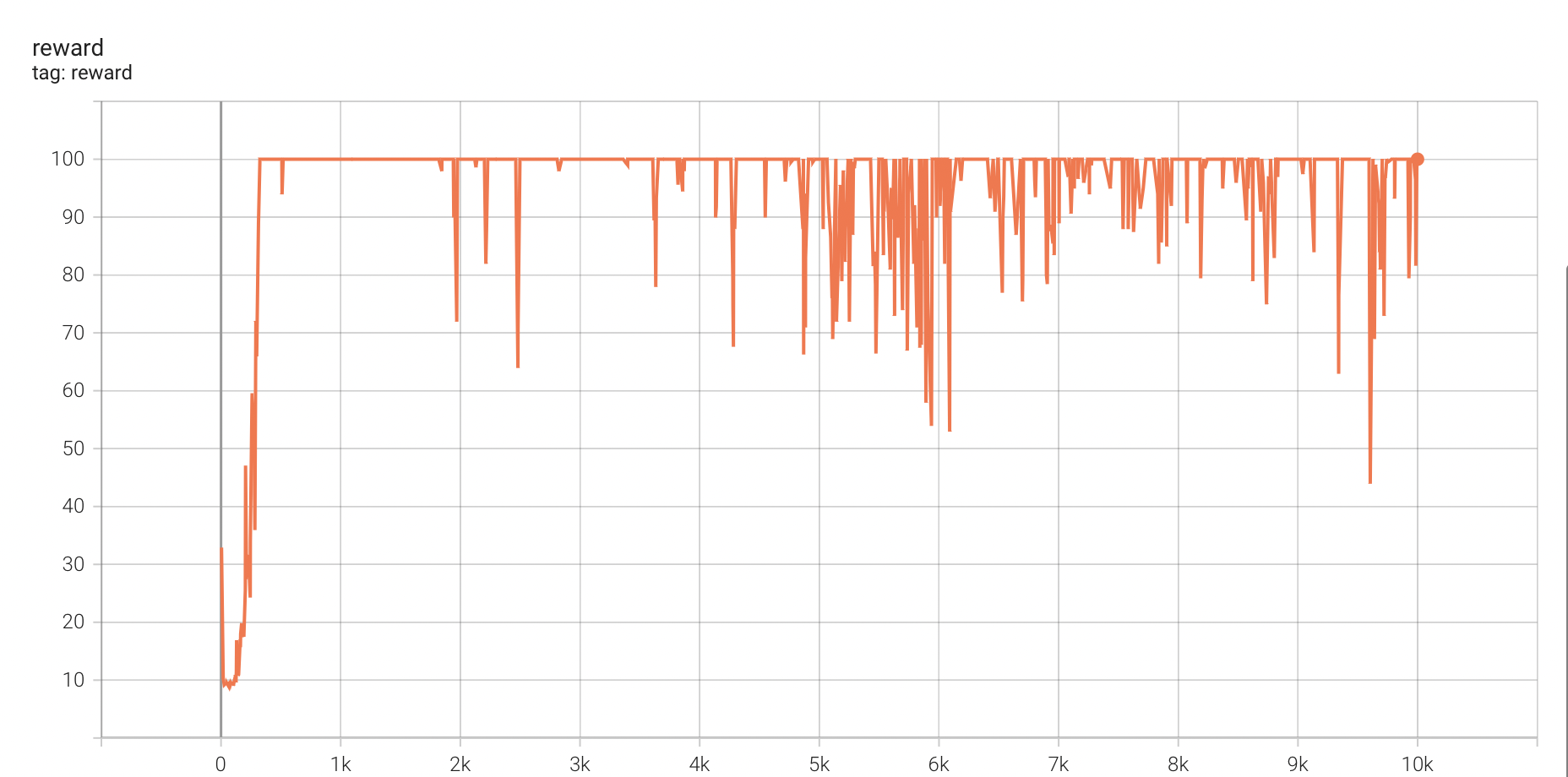}
                \caption{agent6}
                \label{fig:tiger}
        \end{subfigure}%
        \begin{subfigure}[b]{0.125\textwidth}
                \includegraphics[width=\linewidth]{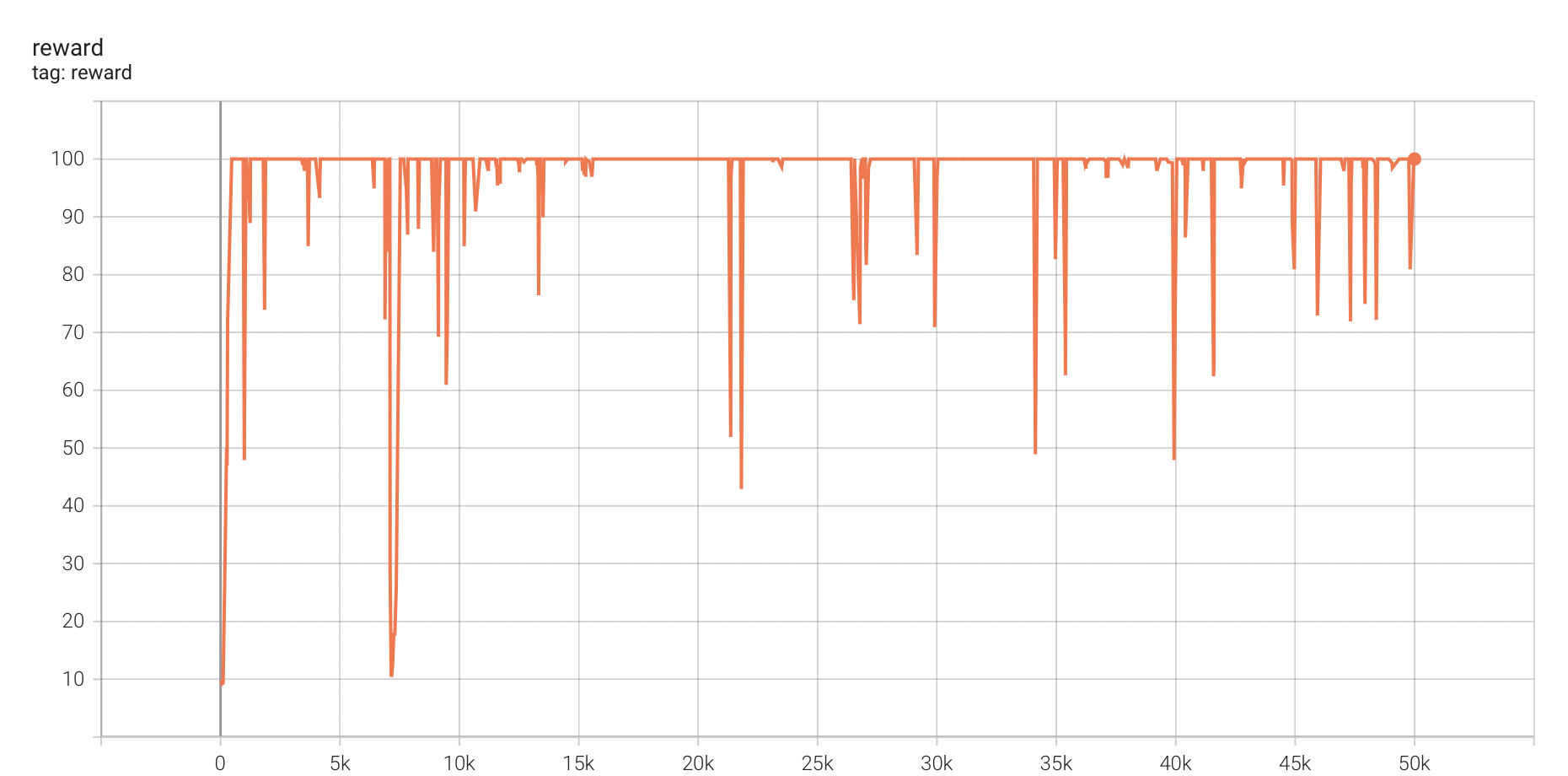}
                \caption{agent1}
        \end{subfigure}%
        \begin{subfigure}[b]{0.125\textwidth}
                \includegraphics[width=\linewidth]{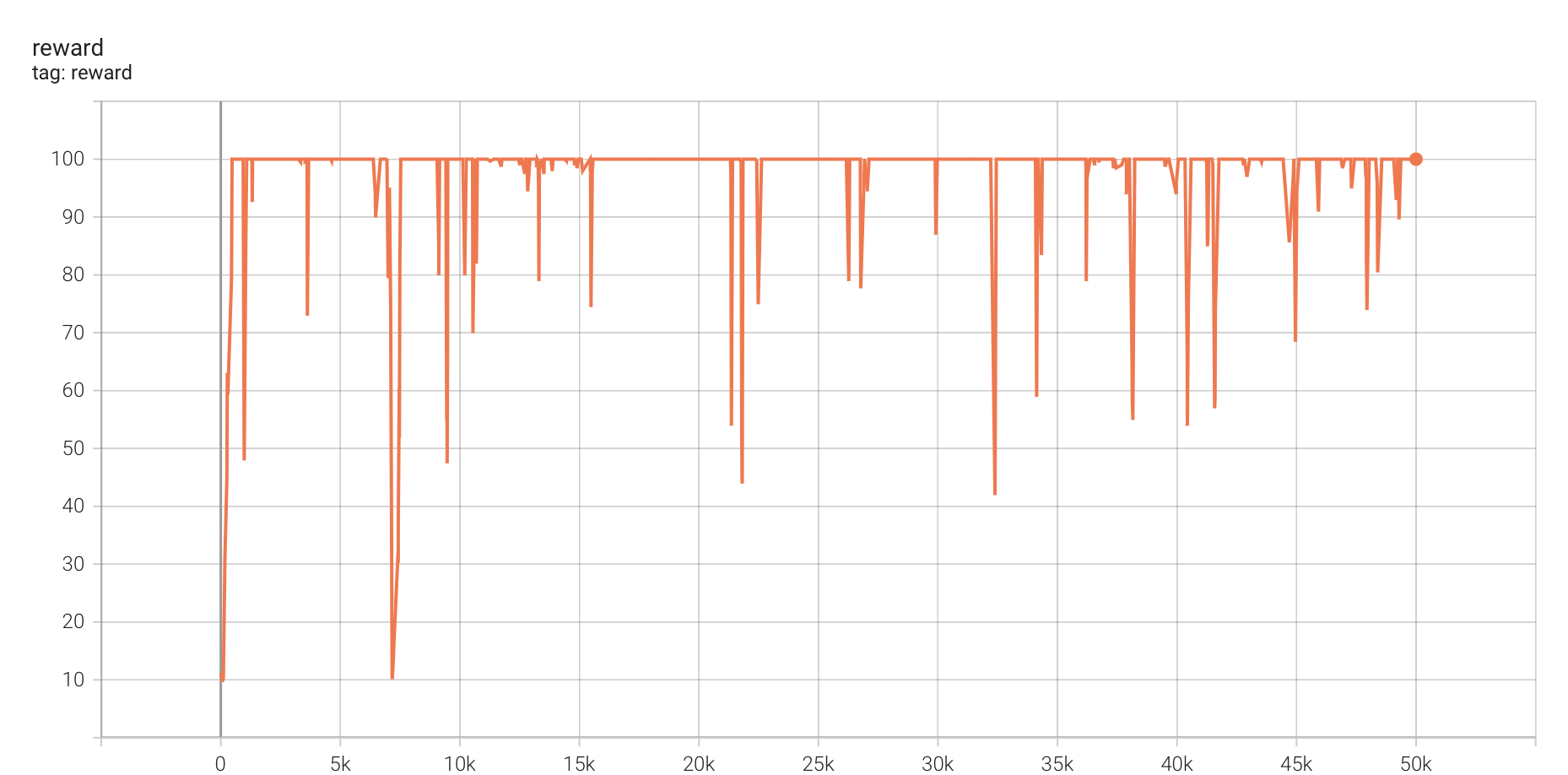}
                \caption{agent2}
        \end{subfigure}%
        \caption{Synchronous control: 6 agents(a-f), 2 agents(g-h)}\label{fig:ddp6agents}
\end{figure*}

We test DDA3C on the task of CartPole-v0 game in OpenAI Gym. In each epoch, we run one episode of CartPole-v0 with a limitation of maximum 100 steps. CartPole-v0 environment will give a reward of +1 for every timestep that the pole remains upright and end when terminal states reached. Hence a total reward of 100 means an optimal policy for a 100-step episode -- every move scores. The epoch minibatch size is set to 100. $k$ is set to 120.

In Figure \ref{fig:animals}, we do the training for totally 50k epochs (the x-axis is the epoch number) and start the knowledge sharing at 20k-th epoch (threshold) for the two-agent group learning case. Every point of the first model update involving shared knowledge in Figure \ref{fig:animals}, \ref{fig:4agents}, \ref{fig:6agents} is marked by a dark line. Figure \ref{fig:1} is the single-agent baseline, in which we can see that the total reward keeps having big fluctuations over the entire training process, namely there are quite a few episodes where bad actions were chosen. The training was unstable and never managed to converge to a stable optimal policy that can choose good actions all the time. In contrast, Figure \ref{fig:2} and Figure \ref{fig:3} keep very stable at 100 after knowledge sharing starts at 20k-th epoch, showing that two-agent group learning quickly managed to maintain a stable optimal policy.

For the game of CartPole-v0, group learning with two agents already has very good performance. We perform more experiments to test DDA3C's ability to scale to more agents. With more agents, we have more experiences that are diversified and the influence of each single early-stage learning mistake can be smaller so that we made attempts to start knowledge sharing earlier. For 4 agent case in Figure \ref{fig:4agents}, we train for totally 20k epochs and start sharing at 10k-th epoch, and for 6 agent case in Figure \ref{fig:6agents} we train for totally 10k epochs and sharing starts at 5k-th epoch. They show that we still reached good results. Agent 4 in the 4-agent group has some very small fluctuations after 10k-th epoch as shown in Figure \ref{fig:tiger1}, meaning that the training stability is near-optimal. All other three agents have converged to stable optimal policy with group learning. As shown in Figure \ref{fig:tiger4}, agent 3 in the 6-agent group was trapped in a bad state before knowledge sharing started and not able to get over it in following studies. It happens sometimes that certain individuals are not doing well, and the probability to see this case can rise when the number of agents in the group goes up. What's interesting is that even in the presence of outliers, others are not affected which shows the robustness of the group learning system. The majority of agents in this 6-agent group works well (5 out of 6) -- stable optimal policy learned after knowledge sharing starts at 5k-th epoch.



\subsection{Comparison with synchronous method}

In Figure \ref{fig:ddp6agents}, we show the reward curve of a 2-"agent" and a 6-"agent" learning system with A2C agent on CartPole-v0 task under synchronous control. This synchronous control is realised through Pytorch DDP (Distributed Data Parallel) library as in DDPPO.  We can see that the curves have severe fluctuations during the entire training period (the more "agents", the more fluctuations), much worse than the single-agent baseline in Figure \ref{fig:1}. As discussed in section \ref{re}, these "agents" are actually worker copies of a single agent and do not form a group learning system. Their learning errors are accumulated during training leading to the worse fluctuations, while in our group-agent learning system the agents took knowledge from peers in an asynchronous manner which gives them the autonomy to utilise knowledge in their own way that could possibly benefit themselves the most.

\section{Conclusion}


GARL describes a very common and important type of real-life learning scenario -- human learning behaviour, which is very promising with regard to application. Any geographically distributed reinforcement learning agents who want to benefit from remote peer learners could join a study group which can be managed by a group agent system. Confirmed by social learning theory \cite{bandura1977social}, rather than being isolated, people learn in a social context with behaviours such as observing and imitating others. This successful learning behaviour of real intelligence should inspire us when we try to create artificial intelligence. 
Joining a group-agent learning system, an agent will be exposed to senior peers and able to learn from them through communication. It provides the agents with the social context that enables them to learn in a more near-human way. There is a body of work adopting social learning \cite{ndousse2021emergent} or imitation learning \cite{stadie2017third,guo2019hybrid} in their agents under single or multi-agent setting. These promising results further reinforce our confidence in the performance that GARL could possibly reach. For future work, we will investigate more sophisticated approaches that work with different types of agents or environments within the group. 

%
%
%
\bibliographystyle{splncs04}
\bibliography{reference}
\end{document}